\documentclass[letterpaper, 10 pt, conference]{ieeeconf}  

\usepackage{epigraph} 
\usepackage{graphicx}  
\usepackage{amsmath}   
\usepackage{subcaption}  
\usepackage{booktabs}      
\usepackage{tabularx}      
\usepackage{array}         
\usepackage{booktabs}      
\usepackage{tabularx}      
\usepackage{array}         
\usepackage[table]{xcolor} 
\usepackage{pifont}        
\usepackage[table]{xcolor} 

\usepackage{pifont}        
\usepackage[hidelinks]{hyperref}

\title{\LARGE \bf
AtomTree: A Hierarchical Framework for State-Aware Embodied
Instruction Following with LLMs
}

\author{
Haotian Zhai\\
University of Minnesota\\
\texttt{haoti002@umn.edu}
\and
Tianming Sha\\
Stony Brook University\\
\texttt{tianming.sha@stonybrook.edu}
\and
Junnan Li\\
University of Wisconsin--Madison\\
\texttt{jli2786@wisc.edu}
}

\begin{document}

\maketitle
\thispagestyle{empty}
\pagestyle{empty}

\begin{abstract}

Embodied instruction following in interactive environments requires agents to interpret natural language commands and complete long-horizon tasks. Existing planners either rely on large labeled trajectories, or use few-shot and zero-shot LLM prompting that produces a whole plan in a single serial reasoning process. These approaches depend on linguistic common sense, commit to a plan from the initial instruction, and only update after failures, which weakens robustness and state awareness. We present AtomTree, a hierarchical framework for state-aware task planning with LLMs. AtomTree incrementally builds a tree of subgoals and actions, where the LLM decides at each step, from the current symbolic state, whether to further decompose or to output a grounded low-level action. A lightweight memory module summarizes past outcomes for reuse in later decisions. When a low-level action stalls, AtomTree triggers hierarchical backtracking replanning to return to the parent node and regenerate the subtree plan from the latest state, preventing continued drift along an incorrect trajectory. In zero-shot evaluation on the ALFRED benchmark, AtomTree outperforms prior LLM-based planners. Project page: \url{https://zhaihaotian.github.io/atomtree-website/}.

\end{abstract}

\section{INTRODUCTION}

\begin{flushright}
\itshape
“Planning is the art and practice of thinking before acting.”\\
--- Patrik Haslum
\end{flushright}

Building embodied agents that can execute diverse household tasks from natural language instructions and rapidly acquire new skills is a core goal in robotics and embodied AI\cite{alfred,li2023behavior,saycan}. Embodied Instruction Following (EIF) requires an agent to jointly perform visual perception, subgoal planning, navigation, and object interaction in interactive 3D environments to complete long-horizon tasks such as ``clean a mug and place it into the coffee machine.''\cite{alfred} In benchmarks such as ALFRED\cite{alfred}, this requires not only language understanding but also continuous instruction decomposition and execution under partial observability. Therefore, how to achieve EIF that is both generalizable and capable of long-horizon planning under limited supervision remains an open problem\cite{flare}.

\begin{figure}[t]
    \centering
    \includegraphics[width=\linewidth]{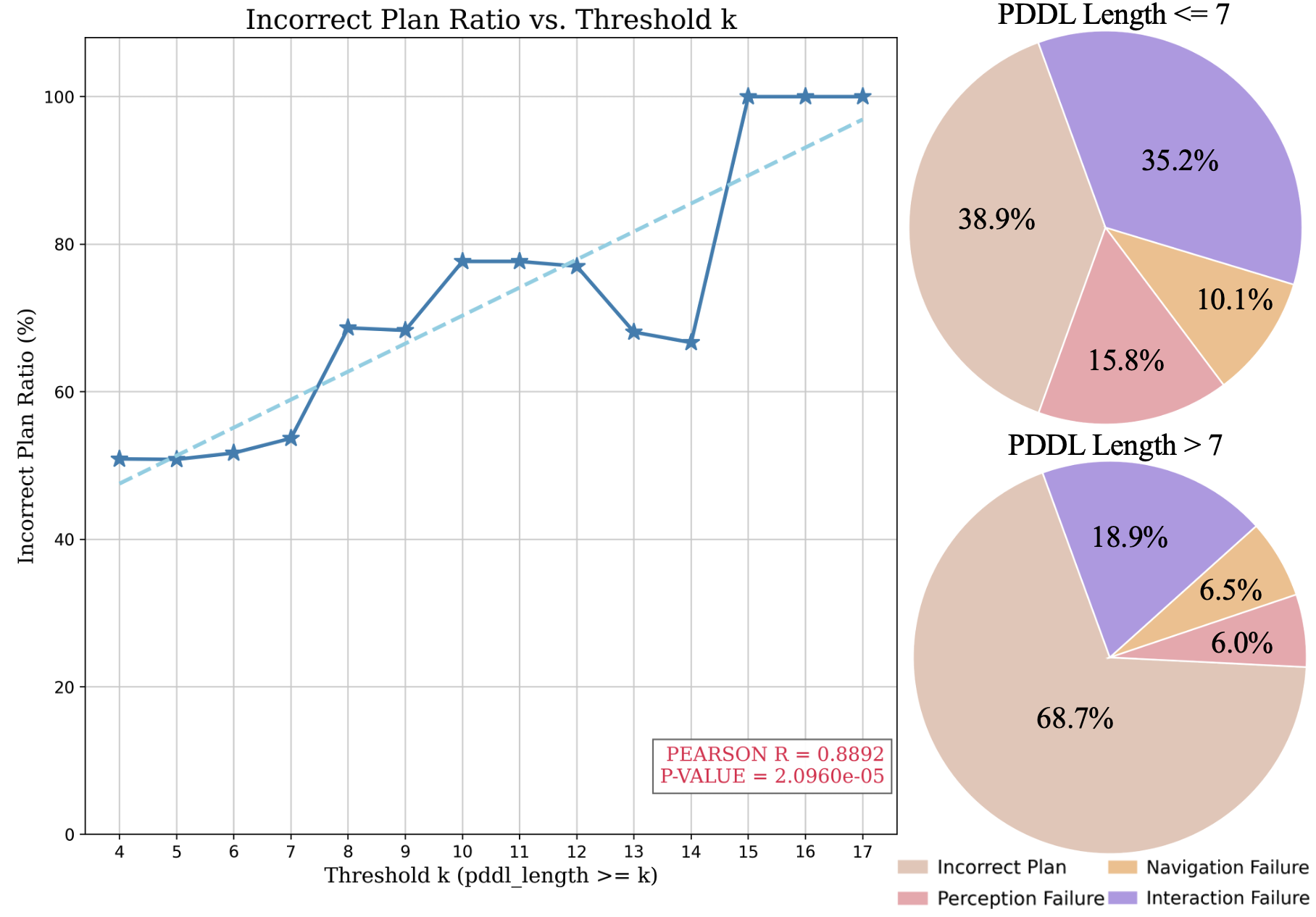}
        \caption{Error analysis on the Valid Seen split of the ALFRED benchmark with respect to ground-truth PDDL plan length, showing that for FLARE, a state-of-the-art few-shot method, incorrect-plan errors dominate overall and become the primary failure mode for long-horizon tasks (PDDL length $> 7$).}

    \label{fig:incorrect-plan-length}
\end{figure}

A straightforward approach is to train policy networks with large-scale pairs of natural-language instructions and expert action sequences\cite{zhao2024epo,murray2022following,pashevich2021episodic,blukis2022persistent,zhang2021hierarchical}. However, both instruction annotation and expert demonstration collection are expensive and time-consuming, and collecting sufficient data in complex environments is often impractical. To reduce annotation cost, several recent studies manually design action sequences for specific task types, but such predefined procedures do not scale well across heterogeneous tasks\cite{min2021film,inoue2022prompter}. Another line of work leverages prior knowledge from large language models (LLMs) for task planning. LLM-Planner\cite{song2023llm} uses few-shot in-context learning to generate high-level subgoal sequences, which are then executed by low-level controllers. FLARE\cite{flare} further introduces visual information during planning. Socratic Planner\cite{socratic} decomposes instructions via self-questioning and performs Socratic-style replanning with visual feedback after failures.

Despite these advances, most LLM-based methods remain unstable on complex long-horizon tasks. On ALFRED Valid Seen, we perform an error analysis of state-of-the-art few-shot planning methods\cite{flare} and categorize failures into four groups in Fig.~\ref{fig:incorrect-plan-length}: planning errors, navigation failures, perception failures, and interaction failures . When task horizon is measured by ground-truth PDDL plan length, planning errors account for approximately $38.9\%$ when the length is no greater than $7$, close to interaction failures at $35.2\%$. When the length exceeds $7$, planning errors increase to $68.7\%$ and become the dominant failure mode. We further group tasks by the PDDL length threshold $k$. As $k$ increases from $5$ to $15$, the proportion of planning errors among all failures rises monotonically from around $50\%$ to nearly $100\%$, showing a strong linear relationship with task length (Pearson $r = 0.8892$, $p \approx 2.1 \times 10^{-5}$). This finding indicates that, in long-horizon embodied instruction following, the primary bottleneck has shifted from low-level navigation, perception, and interaction to high-level task planning itself.

The key limitation of existing LLM-based planners on long-horizon tasks is that decision making is still centered on a single unidirectional sequential plan\cite{alfred,song2023llm,min2021film,inoue2022prompter,socratic}. After receiving an instruction, the model typically generates a full high-level plan in one pass based mainly on linguistic commonsense, and replanning is triggered passively only after repeated subgoal failures, rather than through closed-loop decisions at every step conditioned on the current environment state. Consider the ALFRED instruction ``Put a washed potato slice on the white table.'' The environment ontology does not contain a category named ``white table,'' and only exposes a limited set of semantic classes such as ``Dining Table.'' The agent must infer during execution that ``white table'' is aligned with the observed dining table instance, revealing the gap between human language and environment ontology. By analogy to travel planning, humans rarely commit all details at the outset. Instead, they first decide high-level steps such as ``go to Paris'' and ``book flights and hotels,'' and then refine details based on actual conditions.

Motivated by this observation, we model EIF as a state-conditioned sequential decision process\cite{puterman2014markov} in which, at each decision node, the agent determines whether to further decompose the current objective or directly execute an atomic action based on the current state and subgoal, resulting in an AtomTree composed of decomposition and execution nodes. This state-driven hierarchical reasoning can mitigate error accumulation caused by one-pass sequential planning in long-horizon settings.

Building on this formulation, we propose AtomTree, which explicitly organizes planning as a tree of decomposition nodes and execution nodes. At each node, the policy conditions only on the current environment state, the current instruction segment, and internal memory to decide whether to decompose into finer subgoals or output an atomic action. For example, for the instruction ``slice a tomato with a knife,'' if the agent is already holding a knife in front of the tomato, AtomTree directly selects the \texttt{slice} action. From an initial state, the system instead decomposes the objective top-down into subgoals such as ``find a knife,'' ``pick up the knife,'' and ``move to the tomato,'' followed by progressive refinement and execution. To realize true state awareness, we maintain memory in each node state to summarize historical observations, enabling consistent context for downstream decisions. When a low-level action stalls for an extended period, we do not perform only local replanning for that action. Instead, we trigger a Hierarchical Backtracking Replanning mechanism that returns to the corresponding parent node and regenerates the subtree plan conditioned on the latest environment state. This enables structured top-down adjustment after local failures, rather than accumulating deviation along an incorrect sequential trajectory.

We conduct a systematic evaluation of \textsc{AtomTree} on the ALFRED benchmark. The results show that, under a fully zero-shot setting, \textsc{AtomTree} surpasses state-of-the-art few-shot LLM-based planners on multiple primary metrics, and even achieves better performance than some few-shot and supervised methods, with especially strong gains on long-horizon tasks and complex instructions. Based on these results, our main contributions are as follows:

\begin{itemize}
    \item We propose \textsc{AtomTree}, a hierarchical tree-structured planning framework for EIF, which explicitly models ``decompose / execute'' decisions as a state-aware Markov process.
    \item We design a memory mechanism to enable the planning process to remain continuously aligned with the current environment state during execution, rather than relying on a one-shot sequential plan.
    \item We validate the effectiveness of the framework through zero-shot experiments on ALFRED. Compared with representative few-shot, zero-shot, and supervised baselines, \textsc{AtomTree} achieves stronger overall performance.
\end{itemize}

\section{Related Work}

\subsection{Embodied Instruction Following}
Embodied Instruction Following (EIF) asks embodied agents to understand and execute natural language instructions in 3D environments, typically requiring long-horizon navigation and object interaction\cite{alfred,dorbala2023can}. Prior end-to-end supervised methods directly map visual observations and instructions to fine-grained actions, but rely heavily on large annotated trajectories or task-specific supervision and thus generalize poorly to new environments\cite{zhao2024epo,murray2022following,pashevich2021episodic,blukis2022persistent,zhang2021hierarchical,kim2023context}. More recent work uses hierarchical or modular planning and treats LLMs as high-level planners that generate subgoals or action sequences, then verify and replan with environment feedback or external modules\cite{lin2025flowplan,flare,song2023llm,socratic,inoue2022prompter,shi2024opex}; however, these approaches still depend on few-shot prompting and mostly perform one-shot, sequential planning with limited use of evolving state and execution history. In contrast, our method abandons a fixed two-layer decomposition and performs dynamic, divide-and-conquer planning: at each step, the model decides based on the current state whether to further decompose subgoals or emit executable actions, enabling more state-aware and adaptive long-horizon planning.

\begin{figure*}[t]
    \centering
    \includegraphics[width=\textwidth]{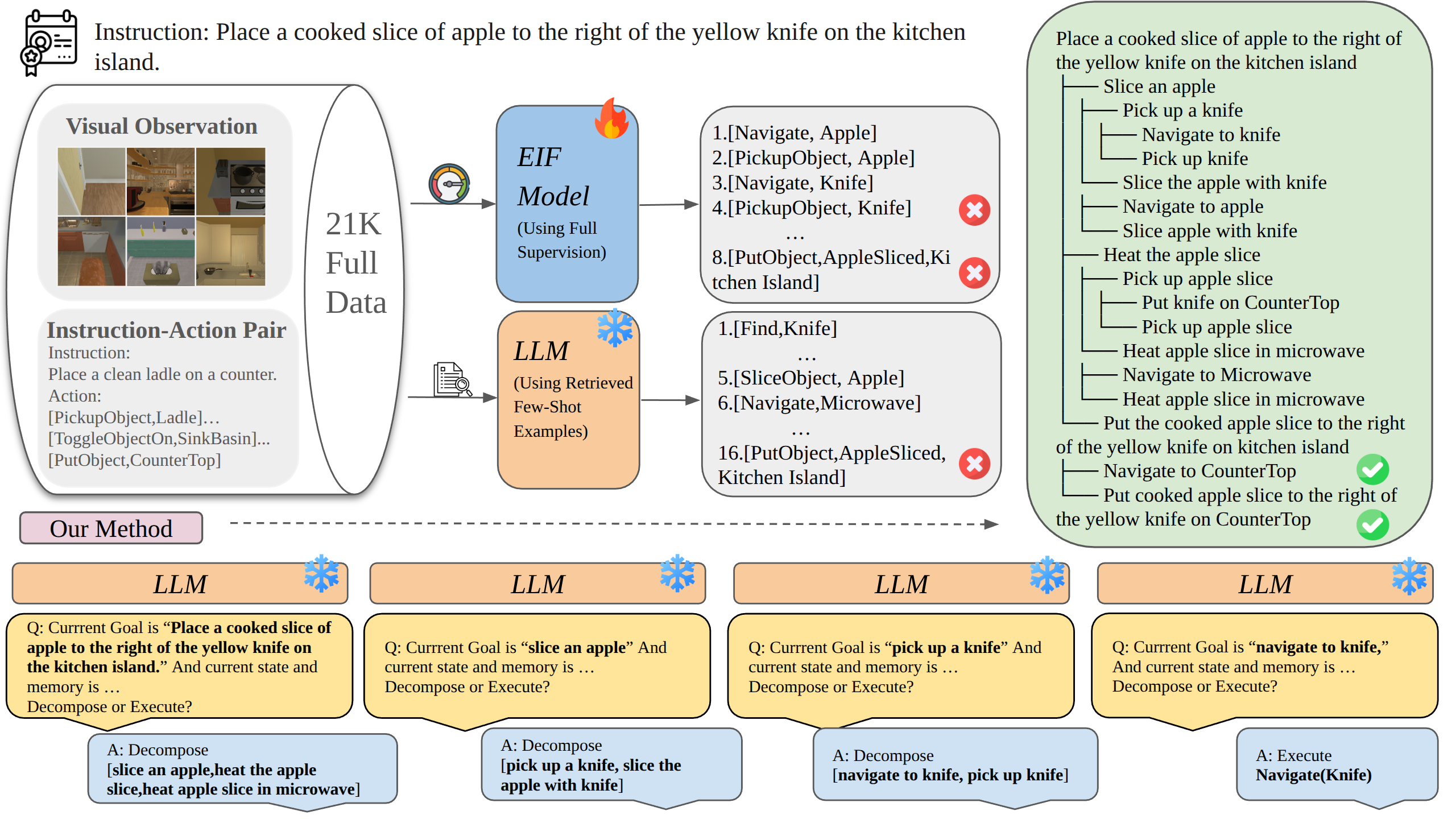}
    \caption{
    Overview of our hierarchical planning framework for embodied instruction following. 
    Given a task instruction and the current visual state, our method recursively decides whether to \emph{decompose} the current goal into finer-grained subgoals or \emph{execute} it directly. 
    This process incrementally builds an explicit hierarchical goal tree, shown in the upper-right panel, where higher-level goals are progressively refined into executable actions. 
    In contrast, we compare against two representative paradigms: (1) a fully supervised end-to-end EIF model trained on the full dataset, and (2) an LLM-based method using retrieved few-shot examples. 
    While these baselines may produce incorrect action sequences, our method maintains a structured tree-based plan and performs step-by-step decomposition conditioned on the current state and memory, enabling more accurate long-horizon reasoning and execution.
    }
    \label{fig:main}
\end{figure*}

\subsection{LLM for Embodied Planning}
With the rapid development of large foundation models (LLMs and VLMs), they have been widely applied to reasoning, planning, and manipulation in robotic systems\cite{driess2023palm,brohan2022rt,brohan2023rt,wake2023chatgpt}. At the task-planning level, a series of works show that decompose-then-solve strategies and structured prompting methods, such as Chain-of-Thought, can significantly improve LLM performance on complex compositional tasks\cite{zhou2022least,wei2022chain}. These results provide a methodological foundation for hierarchical planning in embodied settings.

In embodied planning, many methods follow the paradigm of using an LLM as a high-level planner. Representative approaches include SayCan\cite{saycan}, which combines an LLM with skill affordance value functions, and Inner Monologue–style methods that iteratively revise plans from environment feedback\cite{huang2022inner}. Other works, such as ProgPrompt\cite{singh2023progprompt}, AutoTAMP\cite{chen2024autotamp}, and LLM-State\cite{chen2023llm}, enhance executability and reliability via programmatic prompting, intermediate logical representations, and explicit state tracking. More recent systems for long-horizon tasks\cite{sun2023adaplanner,sun2025retrieval,wong2023learning} adopt bi-stage or hierarchical structures that partially decouple task-level planning from execution-level decision making. However, their planning processes typically still rely on generating relatively long plans in a single shot or on predefined hierarchies, which offers limited support for EIF scenarios like ALFRED that involve long action sequences and highly dynamic states.
We also note the concurrent work ReAcTree~\cite{choi2025reactree}, which is closely related to ours; however, AtomTree is a zero-shot framework that relies only on lightweight symbolic memory, whereas ReAcTree uses explicit control-flow nodes and retrieval-based episodic memory.

\section{Preliminaries}
\label{sec:prelim}

We study embodied instruction following (EIF) on the standard ALFRED benchmark~\cite{alfred}, which is built on top of the AI2-THOR simulator~\cite{kolve2017ai2}. Despite recent progress, the overall task success rate of state-of-the-art methods\cite{flare} on ALFRED remains below 50\%, and our analysis further shows that long-horizon tasks are still the primary bottleneck. In an EIF task, the agent must execute a sequence of actions in the environment to satisfy a language-specified goal. Formally, an EIF task can be represented as a tuple $(I, E, G)$, where $I$ is the instruction, $E$ is the environment, and $G$ is the goal test corresponding to the instruction. In ALFRED, the instruction typically consists of a high-level goal $I_H$ and, optionally, a list of step-by-step instructions $I_L$. In this work, we deliberately discard the step-by-step instruction list $I_L$ and only condition on the high-level goal, following the zero-shot setting in~\cite{socratic} rather than the step-wise language–supervised zero-shot formulation adopted by FlowPlan~\cite{lin2025flowplan}, which better matches real-world human instructions that are often underspecified and rarely spell out every intermediate step.

Following prior work, we adopt a hierarchical planning formulation, but with a key difference: we augment the high-level action space with an explicit \texttt{Decompose} action. At each LLM decision step, the planner conditions on the current state and the current subgoal, and decides whether (1) to further decompose the current objective into finer-grained subgoals, or (2) to directly output an executable high-level atomic action. Concretely, our high-level action space consists of one decomposition action (\texttt{Decompose}), two navigation-related actions (\texttt{Navigation}, \texttt{Find}), and ten interaction actions (\texttt{PickupObject}, \texttt{PutObject}, \texttt{OpenObject}, \texttt{CloseObject}, \texttt{ToggleOnObject}, \texttt{ToggleOffObject}, \texttt{SliceObject}, \texttt{HeatObject}, \texttt{CleanObject}, \texttt{CoolObject}), which is consistent with prior ALFRED-based hierarchical methods~\cite{min2021film,flare,kim2023context}.

The low-level controller follows the standard ALFRED pipeline and maps each high-level action or subgoal to executable primitive actions through a set of deterministic components, including a map-based navigation module and rule-based interaction executors~\cite{blukis2022persistent,min2021film,kim2023context,flare}. In particular, the navigation module performs obstacle-free path planning with deterministic planners such as A* and Fast Marching Method (FMM)~\cite{sethian1996fast}, computing a shortest path or cost-to-go field toward the target and converting the planned route into a sequence of navigation primitives for closed-loop execution. Consequently, once the high-level plan $L_h$ is determined, low-level execution no longer depends on the original instruction $I$, i.e., $P(L_l \mid I, L_h, E) = P(L_l \mid L_h, E)$~\cite{song2023llm}. The low-level controller does not require any additional language-annotated training data.

\section{Method}
\label{sec:method}

We present \textsc{AtomTree}, a state-aware hierarchical planning framework for embodied instruction following in Fig 2. Building on the formulation in Sec.~\ref{sec:prelim}, \textsc{AtomTree} maintains an evolving subgoal tree and interleaves tree expansion with closed-loop execution. Starting from the high-level instruction, the planner iteratively focuses on one active subgoal node, queries an LLM once to obtain a high-level decision conditioned on the current state and memory, and then either expands the subgoal into an ordered list of finer subgoals or executes a single atomic action via the low-level controller.

\begin{figure}[t]
    \centering
    \includegraphics[width=\linewidth]{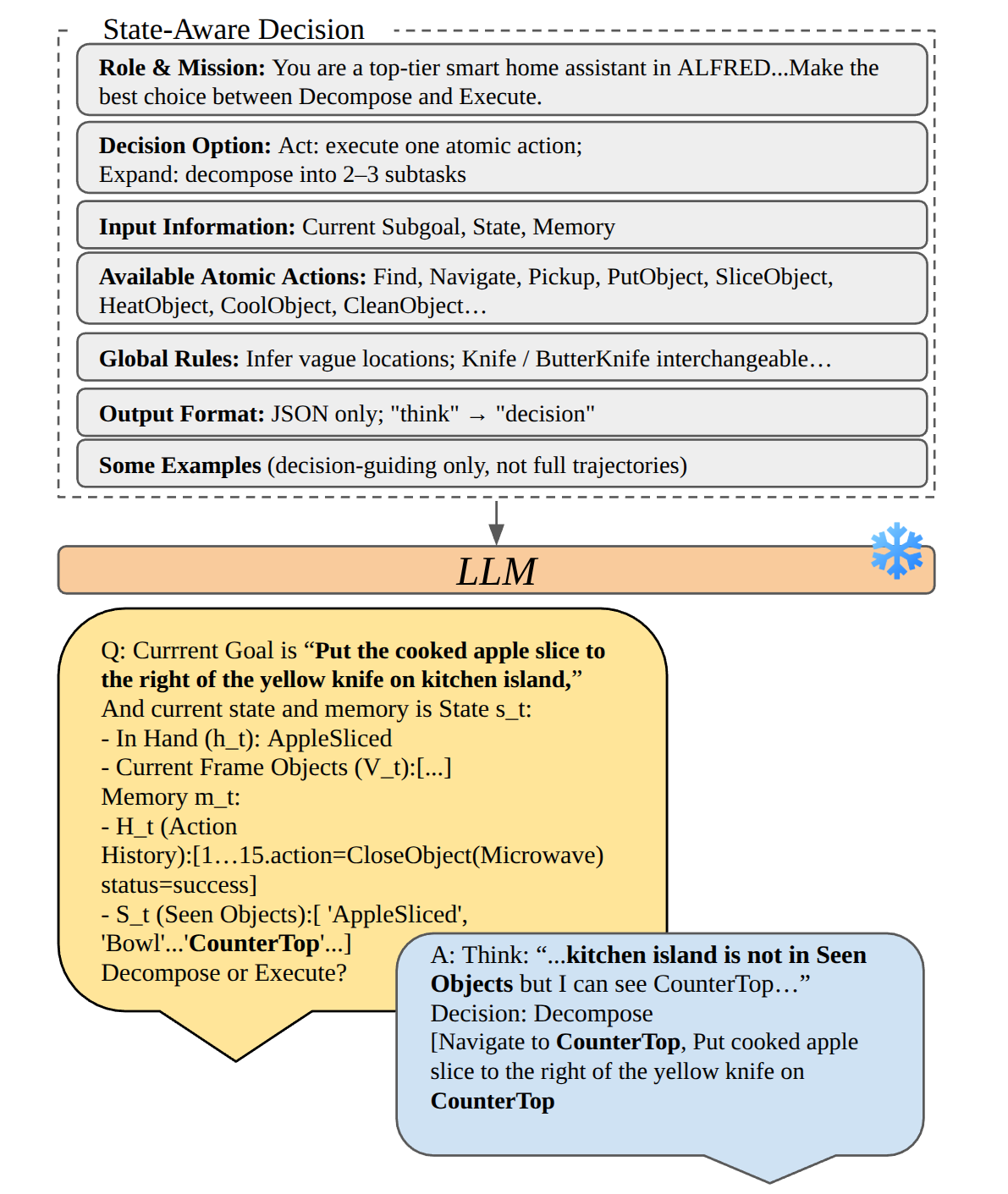}
    \caption{
    Overview of our state-aware decision prompt. The prompt summarizes the role and mission, decision options, input information, available atomic actions, global rules, output format, and a small set of decision-guiding examples. Given the current subgoal together with the current state and memory, the LLM predicts whether to decompose the goal into subtasks or execute it directly.
    }
    \label{fig:prompt}
\end{figure}

\subsection{State-Aware Decision}
\label{sec:method-decompose}

At decision step $t$, the agent maintains a compact environment state
\begin{equation}
s_t = (h_t, V_t),
\end{equation}
where $h_t$ is the object currently held in the agent's hand (or \textsc{None} if empty-handed), and $V_t$ is the set of object names detected in the current egocentric view. The planner also maintains an internal memory $m_t$ (Sec.~\ref{sec:method-memory}).

We represent the planning process as a rooted tree $\mathcal{T}=(\mathcal{V},\mathcal{E})$, where each node $v\in\mathcal{V}$ is associated with a natural-language subgoal $g(v)$. The root node is initialized from the task tuple $(I,E,G)$ with $g(\text{root})$ corresponding to the high-level instruction $I_H$. During execution, the planner repeatedly selects an active node $v$ and queries an LLM-based policy that conditions on the current state, the current subgoal, and memory:
\begin{equation}
y_v \sim \pi(y \mid s_t, g(v), m_t).
\end{equation}

The output $y_v$ takes one of the following two forms.
\paragraph{Decompose.}
The policy may return a decomposition decision together with an ordered list of refined subgoals:
\begin{equation}
y_v = \big(\texttt{Decompose}, [g_1, g_2, \dots, g_K]\big),
\end{equation}
where each $g_k$ is a textual subgoal (e.g., ``find a knife'', ``navigate to the tomato''). We denote the returned list as
\begin{equation}
[g_1, g_2, \dots, g_K] \leftarrow \textsc{DecomposeLLM}(s_t, g(v), m_t),
\end{equation}
and create child nodes $\{v_k\}_{k=1}^K$ with $g(v_k)=g_k$, attaching them under $v$ in $\mathcal{T}$.

\paragraph{Execute.}
Alternatively, the policy may return a single high-level atomic action:
\begin{equation}
y_v = a_v,\qquad a_v \in \mathcal{A},
\end{equation}
where $A$ denotes the set of all high-level actions defined in Sec.~\ref{sec:prelim} other than \texttt{Decompose}.
The selected action $a_v$ is then passed to the low-level controller (Sec.~\ref{sec:prelim}) for closed-loop execution.
After execution, the planner updates $(s_{t+1}, m_{t+1})$ using the new observations and continues with the next active subgoal node. 

\subsection{State-Aware Node Memory}
\label{sec:method-memory}

\textsc{AtomTree} maintains a lightweight symbolic memory that provides long-horizon context under partial observability. At step $t$, the memory is
\begin{equation}
m_t = (S_t, H_t),
\end{equation}
where $S_t$ is the set of seen object names accumulated so far, and $H_t$ is a high-level action history. Each record in $H_t$ stores an atomic action invocation and its final outcome, e.g., \texttt{Find(Apple): 0} and \texttt{Navigation(Table): 1}. In implementation, both $S_t$ and $H_t$ are serialized as short textual lists and included in all subsequent LLM calls.

This memory also bridges the gap between descriptive referring expressions in language and the closed vocabulary of executable entities under partial observability. For example, consider the instruction ``Put a washed potato slice on the white table.'' In ALFRED, the agent can only ground actions on object categories that actually exist in the current scene (e.g., \texttt{DiningTable}, \texttt{SideTable}, \texttt{CoffeeTable}), rather than free-form descriptions such as ``white table.'' At the beginning of an episode, the agent has limited observations and cannot identify which concrete table instance the description refers to. Thus, the planner may initially produce object-centric subgoals that are executable without resolving this reference, e.g., ``slice a potato,'' ``wash the potato slice,'' and ``place the washed potato slice on the table.'' As execution proceeds, the seen-object set $S_t$ is updated (e.g., discovering that the only table in the scene is \texttt{DiningTable}). When revisiting the placement subgoal , the planner can then ground the previously ambiguous reference using $S_t$ and generate an executable action sequence, e.g., \texttt{Navigation(DiningTable)} followed by \texttt{PutObject(PotatoSliced, DiningTable)}.

The same memory interface is used consistently for decision making, decomposition outputs, and backtracking replanning, allowing the policy to exploit discovered objects and prior successes/failures via serialized $(S_t, H_t)$ without passing the full raw interaction trace to the LLM.

\begin{table*}[htbp]
  \centering
  \begin{tabularx}{\textwidth}{llccXXXXXXXX}
  \toprule
  & & & &
  \multicolumn{2}{c}{Valid Seen} &
  \multicolumn{2}{c}{Valid Unseen} &
  \multicolumn{2}{c}{Test Seen} &
  \multicolumn{2}{c}{Test Unseen} \\
  \cmidrule(lr){5-6}\cmidrule(lr){7-8}\cmidrule(lr){9-10}\cmidrule(lr){11-12}
  Method & Venue & Setting & State\&History-Aware
  & SR & GC & SR & GC & SR & GC & SR & GC \\
  \midrule

  \multicolumn{12}{l}{Trained with large-scale instruction--trajectory pairs} \\
  \midrule

  \textcolor{gray}{FILM~\cite{min2021film}} & \textcolor{gray}{ICLR'22} & \textcolor{gray}{Supervised} & \color{red}\ding{56}
  & \textcolor{gray}{$N/A$} & \textcolor{gray}{$N/A$}
  & \textcolor{gray}{$N/A$} & \textcolor{gray}{$N/A$}
  & \textcolor{gray}{25.77} & \textcolor{gray}{36.15}
  & \textcolor{gray}{24.46} & \textcolor{gray}{34.75} \\

  \textcolor{gray}{CAPEAM~\cite{kim2023context}} & \textcolor{gray}{ICCV'23} & \textcolor{gray}{Supervised} & \color{red}\ding{56}
  & \textcolor{gray}{$N/A$} & \textcolor{gray}{$N/A$}
  & \textcolor{gray}{$N/A$} & \textcolor{gray}{$N/A$}
  & \textcolor{gray}{47.36} & \textcolor{gray}{54.38}
  & \textcolor{gray}{43.69} & \textcolor{gray}{54.66} \\

  \textcolor{gray}{MCR-Agent~\cite{bhambri2023multi}} & \textcolor{gray}{AAAI'23} & \textcolor{gray}{Supervised} & \color{red}\ding{56}
  & \textcolor{gray}{$N/A$} & \textcolor{gray}{$N/A$}
  & \textcolor{gray}{$N/A$} & \textcolor{gray}{$N/A$}
  & \textcolor{gray}{30.13} & \textcolor{gray}{$N/A$}
  & \textcolor{gray}{17.04} & \textcolor{gray}{$N/A$} \\

     \textcolor{gray}{Prompter~\cite{inoue2022prompter}} & \textcolor{gray}{IROS'23} & \textcolor{gray}{Supervised} & \color{red}\ding{56}
  & \textcolor{gray}{$N/A$} & \textcolor{gray}{$N/A$}
  & \textcolor{gray}{$N/A$} & \textcolor{gray}{$N/A$}
  & \textcolor{gray}{47.95} & \textcolor{gray}{56.98}
  & \textcolor{gray}{41.53} & \textcolor{gray}{53.69} \\
  

  \midrule
  \multicolumn{12}{l}{Retrieval or prompting with labeled examples} \\
  \midrule

  LLM-Planner~\cite{song2023llm} & ICCV'23 & Few-shot & \color{red}\ding{56}
  & 13.53 & 28.28
  & 12.92 & 25.35
  & 15.33 & 24.57
  & 13.41 & 22.89 \\

  FLARE~\cite{flare} & AAAI'25 & Few-shot & \color{red}\ding{56}
  & $N/A$ & $N/A$ & $N/A$ & $N/A$
  & 31.96 & 41.36
  & 32.57 & 43.23 \\

  \midrule
  \multicolumn{12}{l}{No labeled data required} \\
  \midrule

  LLM-Planner*~\cite{song2023llm} & ICCV'23 & Zero-shot & \color{red}\ding{56}
  & 5.73 & 11.24
  & 4.39 & 12.97
  & 5.71 & 10.42
  & 6.54 & 12.12 \\

  Socratic-Planner~\cite{socratic} & ICRA'25 & Zero-shot & \color{red}\ding{56}
  & 18.78 & 32.20
  & 15.35 & 29.20
  & 18.66 & 30.15
  & 13.47 & 24.89 \\

  \rowcolor{gray!30}
  \textbf{AtomTree (Ours)} & \textbf{$N/A$} & Zero-shot & \color{green}\ding{52}
  & \textbf{40.98} & \textbf{47.02}
  & \textbf{42.39} & \textbf{50.63}
  & \textbf{39.92} & \textbf{46.14}
  & \textbf{41.20} & \textbf{48.32} \\
  \bottomrule
  \end{tabularx}

  \caption{
  Quantitative comparison on ALFRED~\cite{alfred}. We report SR and GC on four splits: Valid Seen, Valid Unseen, Test Seen, and Test Unseen. We annotate each method by its evaluation setting, publication venue, and whether it is explicitly state- and history-aware. For methods that do not report results on certain splits or metrics, we mark them as ``$N/A$''. Results marked with an asterisk (*) are produced by~\cite{socratic}.
  }
  \label{tab:comp-quant-result}
  \vspace{-0.15in}
\end{table*}

\subsection{Hierarchical Backtracking Replanning}
\label{sec:method-replan}

Execution in realistic household environments can fail for many reasons, including interaction errors, target objects being occluded inside containers, or earlier high-level assumptions becoming invalid as new observations arrive. \textsc{AtomTree} addresses these issues with \emph{hierarchical backtracking replanning} at the subgoal-tree level.

For an execution attempt at node $v$, the low-level controller returns an outcome $r_v \in \{0,1\}$ together with an execution status. When an action fails ($r_v=0$) or the controller stalls under its budget\cite{flare,song2023llm}, the planner backtracks to the parent node $u=\text{parent}(v)$ and revisits the higher-level subgoal $g(u)$ under the updated state and memory:
\begin{equation}
y_u \sim \pi(y \mid s_{t'}, g(u), m_{t'}).
\end{equation}
The new decision $y_u$ may again return \texttt{Decompose} with a revised ordered subgoal list, or return an atomic action for immediate execution. The planner then replaces the previously attempted child subtree under $u$ with the newly produced expansion (if any) and resumes execution.

Consider a parent subgoal that requires navigating to a target object, while its current child node first attempts \texttt{Find(Apple)} in order to localize the target but fails ($r_v = 0$). Meanwhile, the accumulated memory records container candidates observed during execution, such as \texttt{Cabinet} or \texttt{Drawer}, in $(S_t', H_t')$. After backtracking to the parent node, the subsequent decision conditioned on the updated state and memory can revise the subtree to include container-oriented search steps, e.g., \texttt{OpenObject(Cabinet)}, and then \texttt{Find(Apple)} again. Once the target is discovered, the planner resumes the original navigation objective. By restructuring the subtree based on newly discovered evidence, \textsc{AtomTree} produces a plan that is better aligned with the current environment state and avoids repeatedly retrying an infeasible action sequence.

\section{Experiments}
\label{sec:experiments}

\subsection{Benchmark and Splits}
\label{sec:exp-benchmark}

We evaluate \textsc{AtomTree} on the standard ALFRED benchmark~\cite{alfred}, an embodied instruction following testbed in interactive 3D household environments built on AI2-THOR~\cite{kolve2017ai2}. We follow the official ALFRED evaluation protocol and report results on both validation and test splits, each consisting of seen and unseen environments. Specifically, the validation set contains {Valid Seen} (820 episodes) and {Valid Unseen} (821 episodes), while the test set contains {Test Seen} (1,533 episodes) and {Test Unseen} (1,529 episodes). ``Seen'' denotes scenes that appear in the ALFRED training set, whereas ``Unseen'' denotes held-out novel scenes for evaluation.

\subsection{Models and Inference Setup}
\label{sec:exp-model}

We use DeepSeek-V3 as the base language model in all experiments. We set the sampling temperature to 1 and cap the maximum number of generated tokens per LLM call to 2,024. Our pipeline operates in a zero-shot manner and does not perform any training on ALFRED trajectories. Although our prompts include a small number of exemplars to constrain the output format, we do not leverage any language-annotated supervision or labeled data for learning.

\subsection{Metrics}
\label{sec:exp-metric}

We report the standard ALFRED metrics \emph{Success Rate} (SR) and \emph{Goal-Conditioned Success} (GC)~\cite{alfred}. SR measures the percentage of episodes in which the agent completes the full task successfully. GC measures the fraction of goal conditions satisfied at the end of an episode, capturing partial progress when a task is not fully completed.

\subsection{Results}

Table~I compares \textsc{AtomTree} with representative supervised, few-shot, and zero-shot methods on ALFRED. We note that some recent methods, such as \textsc{FlowPlan}\cite{lin2025flowplan}, are evaluated under a different setting that requires step-by-step instructions, whereas our method conditions only on the high-level goal instruction. Therefore, in this table, we only report methods that provide results under the goal-only setting. 

Overall, \textsc{AtomTree} achieves the best performance among all zero-shot methods by a large margin on all four splits. Compared with the strongest prior zero-shot baseline, \textsc{Socratic-Planner}\cite{socratic}, our method improves SR from 18.78 to 40.97 on Valid Seen, from 15.35 to 42.39 on Valid Unseen, from 18.66 to 39.92 on Test Seen, and from 13.47 to 41.20 on Test Unseen. The gains on GC are similarly substantial, showing that under hierarchical planning, even when the agent does not fully complete the entire task, it can still satisfy a larger portion of the goal conditions and make meaningful progress toward task completion.

Despite operating in a fully zero-shot setting, \textsc{AtomTree} also outperforms prior few-shot LLM-based planners. In particular, it surpasses \textsc{FLARE}\cite{flare} on both Test Seen and Test Unseen, improving SR from 31.96 to 39.92 and from 32.57 to 41.20, respectively. These results indicate that state-aware hierarchical planning is more effective than one-shot sequential planning even when the latter is assisted by labeled exemplars.

Furthermore, \textsc{AtomTree} is competitive with several supervised methods trained on large-scale instruction--trajectory pairs, and clearly outperforms some of them. For example, our method substantially exceeds \textsc{FILM}\cite{min2021film} and \textsc{MCR-Agent}\cite{bhambri2023multi} on both test splits, while remaining close to stronger supervised methods such as \textsc{Prompter}\cite{inoue2022prompter} and \textsc{CAPEAM}\cite{kim2023context}, especially on unseen scenes. This is notable because \textsc{AtomTree} requires no trajectory training and uses only the goal instruction.

Overall, the results show that \textsc{AtomTree} achieves stronger and more robust embodied planning performance under limited supervision, with particularly clear advantages in unseen and long-horizon settings.

\subsection{Ablation Study}
\label{sec:ablation}

We conduct ablations on the ALFRED validation splits and report SR on Valid Seen and Valid Unseen. We start from a ReAct\cite{yao2022react}-style reactive baseline that re-queries the LLM after each executed atomic action, without hierarchical planning, memory, or backtracking. Since this baseline already performs step-by-step closed-loop decision making, its performance remains close to prior zero-shot sequential planners such as \textsc{Socratic-Planner}.

We then progressively add three components: (i) hierarchical planning with an explicit \texttt{Decompose} action and a subgoal tree, (ii) state\&history memory $m_t=(S_t,H_t)$, and (iii) hierarchical backtracking replanning. As shown in Table~\ref{tab:ablation-atomtree}, each component consistently improves SR on both validation splits.

Hierarchical planning provides the first clear gain by allowing the model to decide whether to further decompose the current objective or directly execute an action. The largest improvement comes from memory, which prevents the planner from forgetting completed actions and falling into repetitive predictions, especially for composite subgoals such as cleaning. Backtracking replanning brings a smaller additional gain by revising the subtree after failures under the updated state and memory.

\begin{table}[t]
  \centering
  \begin{tabular}{lccc cc}
    \toprule
    \# & Hier. & Memory & Replan & Valid Seen SR & Valid Unseen SR \\
    \midrule
    (a) & \color{red}\ding{56} & \color{red}\ding{56} & \color{red}\ding{56} & 15.24 & 14.01 \\
    (b) & \color{green}\ding{52} & \color{red}\ding{56} & \color{red}\ding{56} & 27.20 & 24.73 \\
    (c) & \color{green}\ding{52} & \color{green}\ding{52} & \color{red}\ding{56} & 38.78 & 40.19 \\
    (d) & \color{green}\ding{52} & \color{green}\ding{52} & \color{green}\ding{52} & 40.98 & 42.39 \\
    \bottomrule
  \end{tabular}
  \caption{
  Ablation of \textsc{AtomTree} on ALFRED validation splits. ``Hier.'' enables hierarchical planning with \texttt{Decompose} and a subgoal tree. ``Memory'' adds the state\&history memory $m_t=(S_t,H_t)$. ``Replan'' enables hierarchical backtracking replanning. We report SR on Valid Seen and Valid Unseen.
  }
  \label{tab:ablation-atomtree}
\end{table}

\begin{figure}[t]
    \centering
    \includegraphics[width=\linewidth]{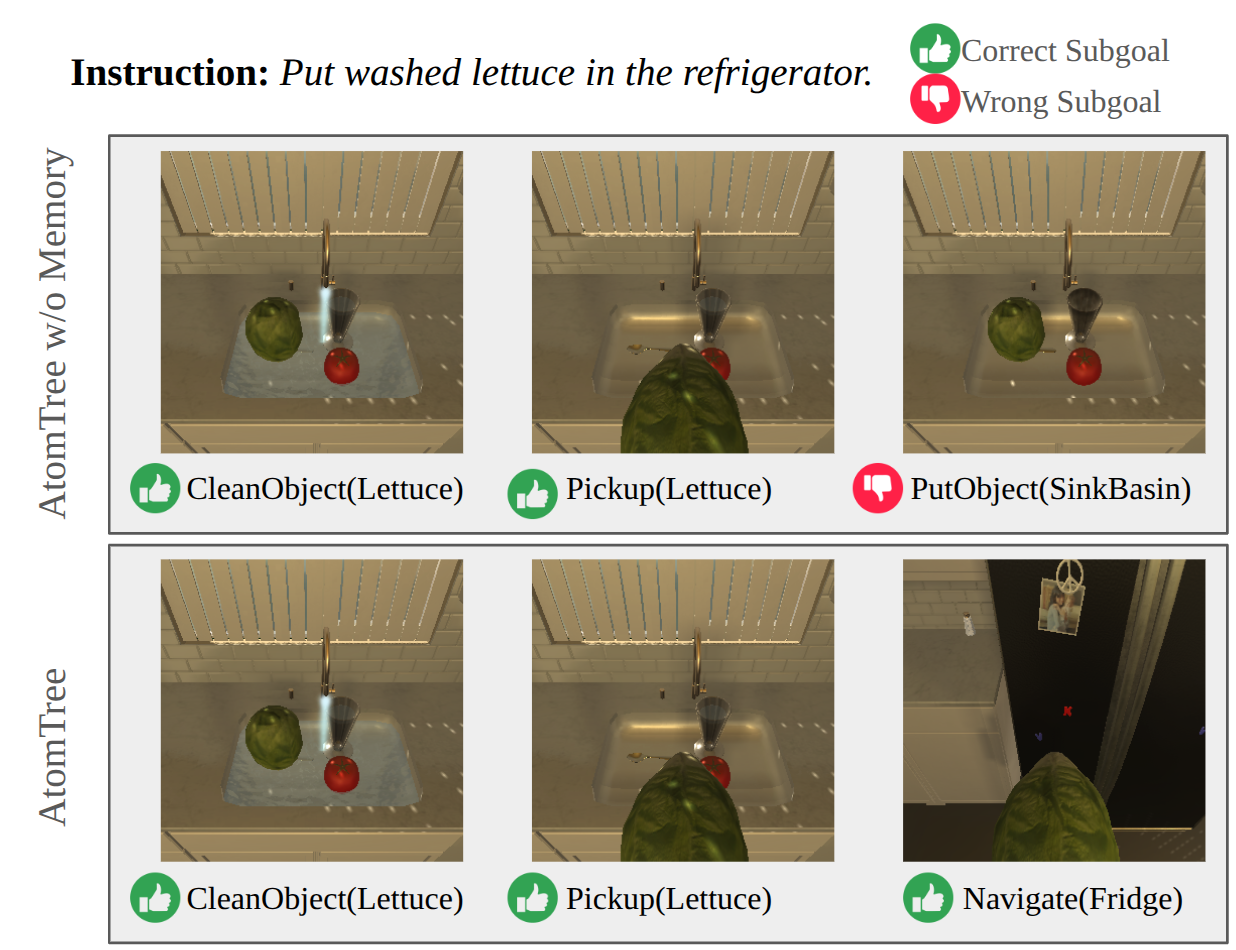}
    \caption{
    Qualitative comparison with and without memory. Without memory, the model predicts an incorrect subgoal after picking up the cleaned lettuce, whereas AtomTree correctly predicts the next subgoal as navigating to the fridge.
    }
    \label{fig:memory}
\end{figure}

\subsection{Qualitative Analysis}

Fig.~\ref{fig:memory} illustrates why memory is important. Consider the instruction of putting a washed lettuce into the refrigerator. A natural decomposition is to first clean the lettuce and then place the cleaned lettuce into the fridge. After the cleaning subgoal has already been completed, the planner should move on to the placement subgoal. However, without memory, the planner cannot reliably tell whether the lettuce has already been washed. As a result, when it receives the next subgoal phrased as ``put the washed lettuce into the refrigerator,'' it may decompose it again into cleaning-related steps followed by placement, leading to redundant subgoals or loops. In contrast, \textsc{AtomTree} records completed actions in memory and correctly proceeds to the next subgoal, namely navigating to the fridge.

\begin{figure}[t]
    \centering
    \includegraphics[width=\linewidth]{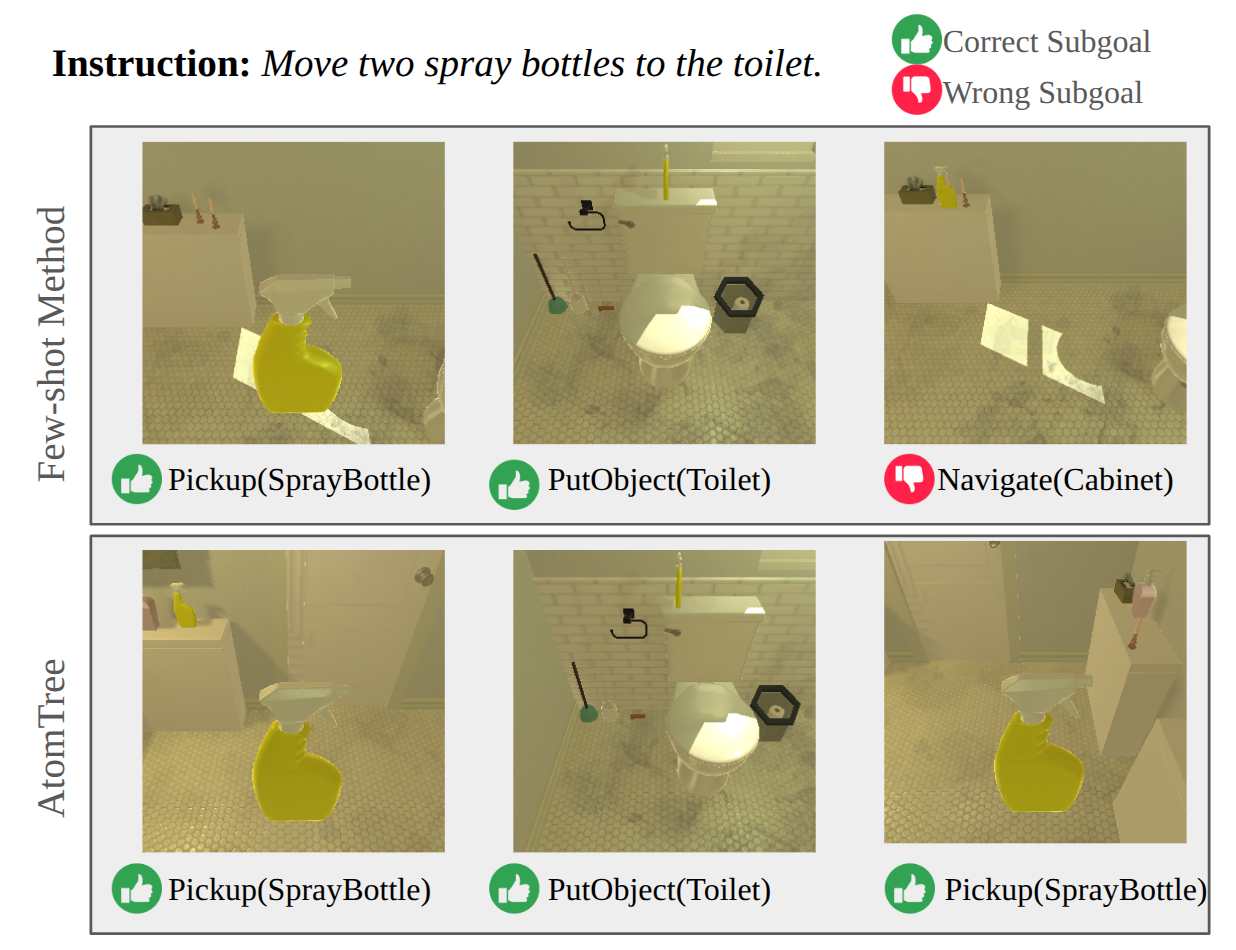}
    \caption{
    Comparison between a few-shot baseline and AtomTree on a multi-instance task. 
    }
    \label{fig:fewshot}
\end{figure}

Fig.~\ref{fig:fewshot} compares \textsc{AtomTree} with a few-shot planner. In this example, after placing the first spray bottle, the few-shot method predicts a cabinet-related action even though the current scene does not require opening a cabinet. This suggests that retrieved exemplars can bias the planner toward irrelevant patterns. In contrast, \textsc{AtomTree} continues with the correct next action by conditioning on the current state and execution history.

\section{Conclusion}

We presented \textsc{AtomTree}, a state-aware hierarchical planning framework for embodied instruction following. Instead of relying on one-shot sequential planning, \textsc{AtomTree} incrementally builds and executes a subgoal tree, using explicit decomposition decisions, lightweight state\&history memory, and hierarchical backtracking replanning. Experiments on ALFRED show that \textsc{AtomTree} substantially outperforms prior zero-shot and few-shot LLM-based planners, and is competitive with several supervised methods, especially on long-horizon and unseen tasks. These results suggest that explicit state-aware hierarchical planning is a promising direction for robust embodied agents under limited supervision.

\bibliographystyle{IEEEbib}
\bibliography{reference}

\end{document}